\documentclass[10pt]{article}
\usepackage[preprint]{tmlr}
\renewcommand{\headrulewidth}{0pt}
\usepackage[utf8]{inputenc}
\usepackage[T1]{fontenc}
\usepackage{amsmath,amssymb}
\usepackage{graphicx}
\usepackage{booktabs}
\usepackage{float}
\usepackage{tabularx}
\usepackage{array}
\usepackage{hyperref}
\usepackage{url}
\DeclareUnicodeCharacter{2212}{\ensuremath{-}}
\DeclareUnicodeCharacter{0394}{\ensuremath{\Delta}}
\newcolumntype{L}[1]{>{\raggedright\arraybackslash}p{#1}}

\title{WorkWorlds: Evaluating Agents in a Company Built Before the Task}

\author{\name Yining Hua \email yininghua@g.harvard.edu \\
      \addr Harvard University
      \AND
      \name Levi Lian \email levilian@stanford.edu \\
      \addr Raycaster \\
      Stanford University}

\def\month{11}
\def\year{2026}
\def\openreview{\url{https://openreview.net/forum?id=XXXX}}

\begin{document}
\maketitle

\begin{abstract}
Many knowledge work benchmarks are built around single tasks, where the context and files for each task are selected concurrently or after task definition. Research has argued that this construction method does not reflect real workplace work, yet the impact has never been measured under controlled conditions. To compare the two constructions under controlled conditions, we introduce WorkWorlds, an infrastructure that modularizes company components into revision, date, and employee seat to be separately adjusted. Using historical data donated by a real company, we reconstructed a pharmaceutical company, PharmaCo. Two experts with over ten years of industry experience each wrote eight tasks based on real events that occurred at the company, and each independently constructed a task-specific environment for every task. Across 192 controlled evaluations, when the agent was switched from the task-specific environment to all the materials truly visible to the employee, evidence access dropped from 90.4 percent to 72.8 percent, and the criterion pass rate dropped from 76.7 percent to 68.0 percent. However, in cases where the agent found the required evidence, the pass rates under both conditions were almost identical. Furthermore, switching between the environments built by the two experts for the same task resulted in an average pass rate difference of 6.6 percentage points. The three agents also gained by different amounts, which changed the gaps between them. We argue that benchmarks should report how environment materials were selected and, if possible, they should test agent score changes after changing the environment builder.
\end{abstract}

\section{Introduction}
Consider the difference between an interviewee solving an interview case and an employee doing their job. What the interviewee gets is the proxy work designed to reflect the actual work. In such a case, the environment is typically designed around the task, and the quality heavily depends on the case designer's field knowledge. In contrast, an employee's work typically arises naturally in the work environment. In this situation, the company's environment, e.g., databases, networks, and histories, exists long before the task, and the employee occupies only a single role within the system, which determines what the employee can see and modify.

Organizational research suggests that these environment elements are inherently part of the work itself -- work is embedded in relationships among people, information, technology, activities, and the organizational context \citep{alter2013}, participants are mutually interdependent \citep{malone1994}, and routines unfold within situated action \citep{feldman2003}. From these perspectives, building an environment for a task takes away important parts of real workplace work. Even when a domain expert builds the environment, the quality of the environment would still depend on the expert's personal experience, which makes it hard to reflect real work. In addition, two experts may build substantially different environments for the same task.

To measure these effects, we design WorkWorlds, an evaluation infrastructure that builds the company before adding tasks. A WorkWorlds company has three parts that can be changed one at a time: revision (which version of the company materials is used), date, and employee seat. The seat and the date together decide what the agent can see and change. We used historical data donated by a real company to reconstruct a pharmaceutical company, PharmaCo, for running the comparative analysis to measure (1) how agent scores change when the environment is built for the task instead of taken from the company; (2) how much the scores change when different experts build the environment for the same task. We hired two experts, each with over ten years of industry experience, to write tasks based on real events at the company, and we merged them into one set of eight tasks across six employee positions based on iterative discussion. We also asked each expert to build a task-specific environment for every task by selecting files from the company, as current benchmarks do.

Through the experiments, we found that when agents worked with all the materials the employee could see instead of the experts' environments, the criterion pass rate dropped from 76.7\% to 68.0\%. Evidence access also dropped from 90.4\% to 72.8\%. Switching between the two experts' environments for the same task also changed the pass rate by 6.6 percentage points on average. We argue that benchmarks should report how environment materials were selected and, if possible, they should test agent score changes after changing the environment builder.

\section{Related Work}\label{sec:related}

\subsection{Placing Multiple Tasks in the Same Workplace}

Several workplace agent benchmarks have started to place multiple tasks in the same company, project, or workspace. TheAgentCompany builds a simulated software company with internal websites, code repositories, chat tools, and simulated coworkers, where agents complete tasks as employees \citep{xu2025theagentcompany}. EnterpriseOps-Gym builds enterprise databases, tools, and policies across eight business domains, with tasks written by experts \citep{malay2026}. Both studies share the same infrastructure across tasks, but each task creates or resets its own environment, so the tasks do not share one company state. In addition, their companies are made of the systems and data the tasks need. They do not have the history of past operations or the different views of different employees that a real company has.

APEX-Agents is the closest to our design. Domain experts first build a project world with files and tools, and then write multiple tasks from it \citep{vidgen2026apex}. However, the world is a project, not a company. It has no employee-specific views or dates, so every task sees the same pool of materials. EnterpriseBench models organizational hierarchies, business data, and role-based access control, so different personas can see different data, and its tasks are generated from this organizational context \citep{vishwakarma2025}. In contrast to WorkWorlds, it has no company versions or dates, and it does not require the same role to see the same environment across tasks.

\subsection{Benchmarks Evaluating Material Availability}

Another line of work asks whether agents should be given curated materials or find the materials themselves. Most benchmarks treat material selection as preprocessing, not as a design choice that needs to be reported and tested. For example, GDPval gives tasks the reference files chosen by the experts who wrote them \citep{patwardhan2026gdpval}, and FORTE gives office tasks across 15 professions their own input files \citep{longcat2026forte}. Both test only the curated setting. Workspace-Bench gives agents large workspaces tied to worker profiles and argues that benchmarks built on pre-specified files leave out much of how workers find and use files \citep{tang2026workspace}. SWE-bench compares giving the relevant files directly with retrieving them from the whole code repository \citep{jimenez2024swebench}, which is the closest to our experiment. However, it tests retriever quality, not environment design, and its repositories are public, with no roles or permissions. Long-context studies also show that irrelevant or distracting context lowers model performance \citep{shi2023distracted,liu2024lost,hsieh2024ruler}, but they use synthetic distractors instead of real materials from a workplace.

\section{WorkWorlds}\label{sec:workworlds}

In contrast to the systems in Section~\ref{sec:related}, which all build their environments on top of tasks, we design WorkWorlds from theories of work and organizations. Work system theory views work as embedded in relationships among participants, information, technology, and activities \citep{alter2013}, with participants mutually interdependent \citep{malone1994}. Organization theory views organizations as structures defined by division of labor and coordination \citep{march1958,mintzberg1980}, where a job determines what a person can see and modify. Routine theory separates lasting organizational routines from each performance of them at a given point in time \citep{feldman2003}.

From these theories, a company has at least three attributes: (1) employees working inside shared environments, (2) jobs and permissions that limit what each person sees and edits, and (3) a history of past events from daily operations (so every evaluation sits at some point in time). We build WorkWorlds around these attributes so that it is still a complete workplace even with no tasks added.

\begin{figure}[t]
\centering
\includegraphics[width=\textwidth]{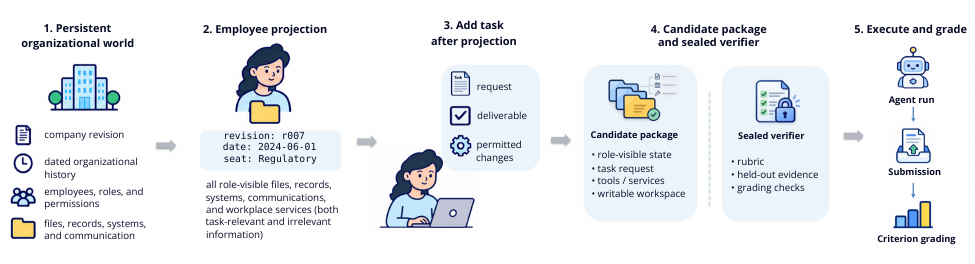}
\caption{Overview of WorkWorlds. The revision, date, and employee seat create an employee projection from a persistent company world. The task is introduced only after this projection is fixed. Information visible to the candidate is then kept apart from the scoring information on the verifier side, and the agent runs the task and is scored.}
\label{fig:overview}
\end{figure}

For each evaluation, we fix a value for each of them, and we can vary one while keeping the others the same: a revision (which version of company materials to use), an employee seat (the agent's job and permissions), and a date (which past events already happened). We define the employee projection as everything the agent can see under these values, which includes files unrelated to the task. The task is constructed only after the projection is set (Figure~\ref{fig:overview}).

\subsection{PharmaCo}

We implement PharmaCo based on WorkWorlds using materials donated by a real company covering regulatory, manufacturing, quality, supplier quality, and stability work. All materials were de-identified and rewritten during preprocessing so that none of the records match a real product, filing, or person. For this study we construct six employee seats. Each seat sees hundreds of files (775 for the Manufacturing Science and Technology (MSAT) and Quality Control (QC) seats), most of them on shared site drives, while personal or confidential materials are only open to certain job roles and owners. Only a small share of these files matters for any one of the tasks. Appendix~\ref{app:B} has more details on PharmaCo's organization, documents, and permissions.

Every task in PharmaCo is an event that actually happened in the company's history. We hired two domain experts, each with over ten years of industry experience, to write these events into tasks separately, giving each task an employee seat and a date, and then merged the two versions into one set through iterative discussion between the two experts. The rubrics were written in reference to the outcomes of real events, and the same experts reviewed every criterion to revise anything that did not reflect how professionals in the role would actually do the work. We also ran a few exploratory evaluations first to refine the criteria. All criteria score whether the deliverable content is ready for sign-off. All criteria are deterministic or source-based semantic checks, linked to relevant evidence.

\subsection{Implementation}

We implemented WorkWorlds on Harbor \citep{harbor2026}. All worlds share the same evaluation code, and each world only adds its own code for building the company, so a new company can be added without changing the evaluation code. Company files are generated from structured world data, so the same revision always gives the same files. Interactive systems (e.g., record systems and internal messages) are built as Model Context Protocol (MCP) tools \citep{anthropic2024mcp}. They use the same employee identity as the file system, so the agent's permissions do not change with the entry point it uses.

To make sure the company is the same every time, we record the checksum of every generated file in a list and check each rebuilt company and each employee workspace against this list before a run. Some company systems (e.g., record systems) can be changed by the agent during a task. For these, we only allow the changes the agent's role permits and log every change in order, and replaying the log afterward shows if the final company state is correct. Appendix~\ref{app:A} gives the construction rules, and Appendix~\ref{app:D} describes the checks.

\section{Experiments}\label{sec:experiments}

\begin{figure}[t]
\centering
\includegraphics[width=0.92\textwidth]{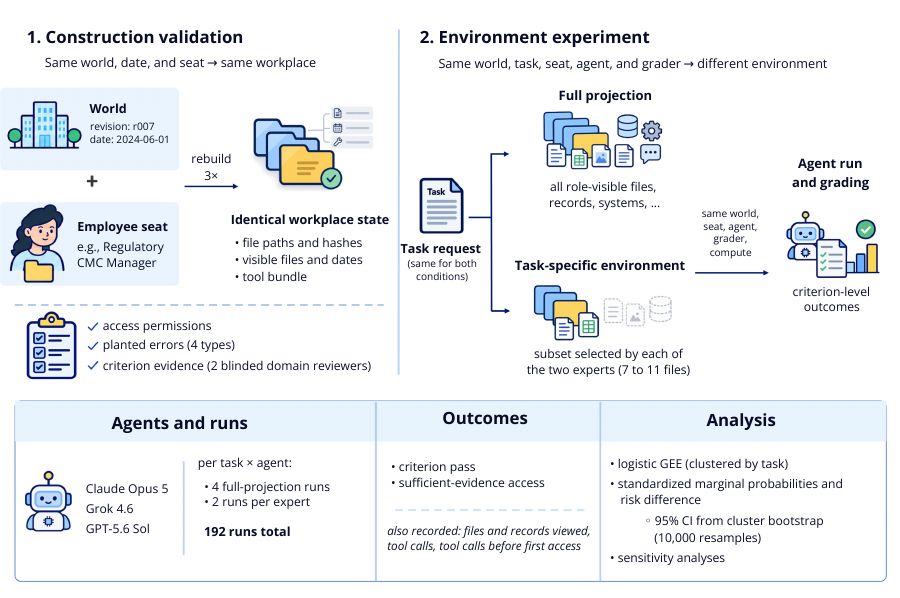}
\caption{Experimental design. Construction validation tests if a fixed revision, date, and employee seat reproduce the same workplace. The environment experiment changes only the materials the agent receives.}
\label{fig:design}
\end{figure}

We ran two experiments on the eight PharmaCo tasks (Figure~\ref{fig:design}): (1) construction validation, which asks if each employee projection holds exactly what the employee could see, so that the full projection in the second experiment is what the employee could actually see; (2) the environment experiment, which changes only the materials handed to the agent and looks at how performance changes.

The eight tasks (stability update, overwrap position, validation defense, authority response, records integrity, supplier-change audit, shelf-life assessment, and qualification sweep) have 72 criteria in total. They all use one revision and one date across six employee seats, with two pairs of tasks sharing a seat (Appendix~\ref{app:B}).

\subsection{Construction Validation}

To validate construction, we rebuilt PharmaCo three times and created every seat three times, and all file paths and SHA-256 digests had to match across copies. Tasks sharing a seat had to start from the same workspace, and each task's loaded workspace had to match its seat. A permission audit then looked for personal, cross-department, or verifier-only materials that did not belong. On top of this, we planted four errors (a modified file, an unauthorized personal file, a renamed verifier file, and a deleted evidence source), which the checks needed to catch before any run. Finally, two other domain reviewers, blind to model outputs and expert environments, judged if each criterion had enough evidence inside the employee's view, and we filled any gap before locking the evidence map.

Three more worlds, rebuilt from public materials of an oncology drug company, a diagnostic device company, and a clinical site (Appendix~\ref{app:B}), went through the same checks, as did the public sample world (12 conformance checks and six planted isolation violations). These worlds serve only to show that the construction method carries over to other companies.

\subsection{Environment Experiment}

For the environment experiment, each run gives the agent either the full projection, meaning everything the seat can see on the task date, or a task-specific environment, meaning the materials one expert picked while writing the task, as current benchmarks do. Experts could only pick from existing materials, without adding, summarizing, renaming, or reorganizing anything. Nothing else differs between the two conditions (world, task, seat, date, rubric, grader, agent settings, and compute budget).

We tested Claude Opus 5, Grok 4.6, and GPT-5.6 Sol, with four full-projection runs and two runs per expert environment for each task and agent (192 runs), interleaving the two conditions over the same period. Since merging changed the tasks and rubrics a little, every expert environment was rechecked for enough evidence before the runs. Two of the second expert's environments fell short, and the expert added one file to each (Table~\ref{tab:E4}). Once all runs finished, a GPT-5.6 Sol grader scored the semantic criteria in one batch, blind to condition and environment, and domain experts rechecked 240 of its scores by hand (Appendix~\ref{app:E}).

\subsection{Outcomes and Statistical Analysis}

We define a criterion execution as one criterion scored in one run, and we set two outcomes in advance. We define criterion pass as whether the criterion execution passes, and sufficient-evidence access as whether the agent actually read one of the evidence sets listed for that criterion (seeing a filename or a directory listing does not count; Appendix~\ref{app:E}). Traces that cannot show whether the evidence was read get an unknown label, and we report results both with every unknown counted as accessed and with every unknown counted as not accessed. Passing without reading the listed evidence is still possible, because other correct paths may exist. The pass rate after finding evidence is only descriptive, because access itself differs between conditions. Per run, we also count files and records viewed, tool calls, and tool calls before the evidence was first found.

Because criteria within a task share one request and one environment, they are not independent. We fit a logistic generalized estimating equation (GEE; \citealp{liang1986}) with condition and agent as covariates, task as the cluster, and an exchangeable working correlation. GEE fits our target, which is the average difference between conditions across tasks, whereas a mixed-effects model would need a task-level variance estimated from only eight tasks \citep{hubbard2010}.

Standardized probabilities come from predicting every observed criterion execution under both conditions and averaging, and their difference is the reported risk difference. Access is analyzed the same way on executions with an observable label. Intervals (95\%) come from resampling whole tasks 10,000 times and refitting the model, and with only eight tasks their real coverage may be off from 95\%. As sensitivity analyses, we used equal task weights, left out one task at a time, and estimated effects separately by agent and by expert.

\section{Results}\label{sec:results}

\subsection{A Company Supporting Multiple Evaluation Instances}

Results show that WorkWorlds built the same company every time. All three rebuilds of PharmaCo were identical, and each of the 18 seat instantiations gave back its own state (six seats, six distinct fingerprints). Tasks that share a seat also started from the same projection, and the workspace loaded for each task matched its seat.

We also found no unauthorized exposure in the permission audit, and all 1,597 expected files were in place. Only one file appeared on both the candidate and verifier sides, which was a protocol the employee is supposed to see and the grader also cites (Appendix~\ref{app:D}). The checks caught all four planted errors before any run.

For evidence sufficiency, the two reviewers agreed on 66 of 72 criteria, and the evidence they picked largely overlapped (mean Jaccard 0.77; Appendix~\ref{app:D}). Seven criteria went to adjudication. One of them was a real gap in the environment, and we fixed it.

Outside PharmaCo, the same construction also held. All 12 conformance checks passed in the public sample world, and all six planted violations were rejected. For example, a quality employee could not see an approval record before its effective date, but could see it after. The three secondary worlds were built with the same rules (Appendices~\ref{app:B} and~\ref{app:D}).

\subsection{Building Environments Around Tasks Changes Benchmark Scores}

All 192 runs were scored with no grader errors, which gave 1,728 criterion executions (864 per condition; failed attempts were rerun, Appendix~\ref{app:E}). The grader and domain experts agreed on 93.8\% of 240 sampled executions (Cohen's kappa = 0.85), and agreement was about the same in both conditions (92.5\% and 95.0\%). For access labels, two independent coders agreed on 89.6\% of 346 sampled executions (Cohen's kappa = 0.73).

\begin{figure}[t]
\centering
\includegraphics[width=\textwidth]{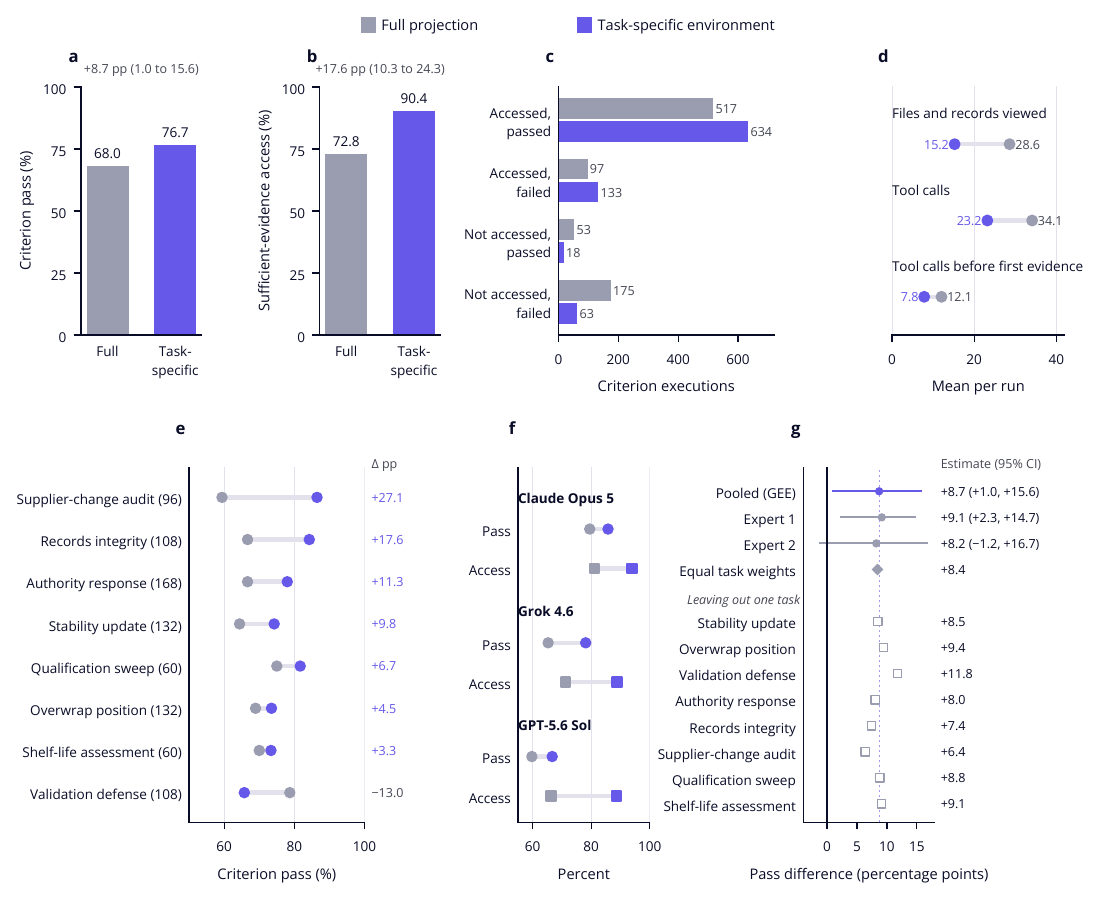}
\caption{Effects and consistency of the task-specific environment across 72 criteria. The top row shows criterion pass, sufficient-evidence access, the joint state of access and pass, and mean search volume per run. The bottom row shows criterion pass by task, criterion pass and evidence access by agent, and the sensitivity of the pooled pass difference to expert, task weighting, and leaving one task out. Criterion pass and sufficient-evidence access are GEE-standardized marginal probabilities. Differences are the task-specific environment minus the full projection. Where shown, intervals are 95\% task-clustered bootstrap intervals.}
\label{fig:results}
\end{figure}

We found that agents scored higher when they were given a task-specific environment (Figure~\ref{fig:results}, top left). The criterion pass rate went from 68.0\% under the full projection to 76.7\%, a difference of 8.7 percentage points (95\% interval 1.0 to 15.6), which was statistically significant. Evidence access changed even more, from 72.8\% to 90.4\% (17.6 percentage points; 10.3 to 24.3). Unknown access labels were rare (2.5\% and 1.9\% of executions), and even when we coded them in the most extreme way, the access difference stayed between 15.2 and 19.6 percentage points (Appendix~\ref{app:E}).

Interestingly, agents that found the evidence did about as well in both conditions (84.2\% passed under the full projection and 82.7\% under the task-specific environment). What differed was how often they found it. Under the full projection, 175 criterion executions failed because the agent never reached the evidence, compared with only 63 under the task-specific environment (Figure~\ref{fig:results}, top center). We also asked what the full projection would score if agents had found the evidence as often as in the task-specific environment, keeping their own pass rates with and without evidence. The answer was 78.4\%, which is a bit higher than the 76.9\% observed in the task-specific environment (raw counts, Table~\ref{tab:E1}). In other words, the gap in access explains all of the gap in pass rates.

Agents also had to search more under the full projection (Figure~\ref{fig:results}, top right). They read files and records 28.6 times per run on average, compared with 15.2 in the task-specific environment (a file read twice counts twice). They also needed 12.1 tool calls to first reach the evidence, compared with 7.8. The three agents all used about 10 to 12 more tool calls (Opus 5: 30.5 vs. 20.7; Grok 4.6: 33.8 vs. 23.2; GPT-5.6 Sol: 38.1 vs. 25.8) under the full projection.

Across tasks, pass rates were higher in the task-specific environment for seven of the eight tasks (Figure~\ref{fig:results}, bottom row). The largest difference was in supplier-change audit (27.1 percentage points), which also had the largest gain in access (35.2). In contrast, validation defense went the other way. Its pass rate was 13.0 percentage points lower with both experts' environments, although agents found the evidence equally often in the two conditions (78.5\%).

The result held when we left out any one task (pass differences from 6.4 to 11.8 percentage points) and when we weighted tasks equally (8.4). All three agents improved, and their overall ranking did not change (Claude Opus 5 from 79.5\% to 85.8\%, Grok 4.6 from 65.3\% to 78.1\%, and GPT-5.6 Sol from 59.7\% to 66.7\%). However, Grok 4.6 gained about twice as much as the other two, so its gap to Claude Opus 5 shrank from 14.2 to 7.7 percentage points. Within single tasks, the best agent was a different one in three of the eight tasks (stability update, authority response, and shelf-life assessment), although each comparison only rests on 20 to 56 criterion executions per agent and condition.

The experts' environments were very small. They had a median of nine materials (range 7 to 11; Table~\ref{tab:E4}), which for the four MSAT and QC tasks was only 0.90\% to 1.29\% of the 775 files the employee could see. The two experts also did not pick the same materials in five of the eight tasks (mean Jaccard 0.87).

Overall, the two experts' environments raised the pass rate by similar amounts (+9.1 and +8.2 percentage points). Within the same task, however, switching from one expert's environment to the other's changed the pass rate by 6.6 percentage points on average (95\% interval 4.2 to 9.2; 1.2 to 13.3 across tasks). This is about one third of the 19.8-point gap between the strongest and weakest agents under the full projection. Three environments even scored lower than the full projection: both experts' environments for validation defense ($-$8.3 and $-$17.6) and the first expert's environment for shelf-life assessment ($-$3.3; Appendix~\ref{app:E}).

\section{Discussion and Limitations}\label{sec:discussion}

Workplace benchmarks often build an environment for each of the tasks. No one has ever tried to measure how much this choice of building environments for tasks makes the score deviate from actual work. In this study we propose WorkWorlds, which builds the company first and then runs the same tasks under two different environments: actual environments that exist before the task vs. environments experts selected for the tasks.

We found that changing the environment for the same agent doing the same task changed the score a lot. In cases where the agents were able to identify the evidence, their pass rates were similar under the two different settings. This partially disagrees with long-context studies \citep{shi2023distracted,liu2024lost,hsieh2024ruler}, where people found that the existence of extra context would lower agent performance even when the needed information exists. In our experiments, extra materials don't hurt agent performance as long as the agent found the necessary evidence. However, validation defense went the other way---in this task, the agent found the evidence equally often in expert-constructed environments; and when it found the evidence, the pass rate was even lower (78.6\% vs. 92.9\% under the full projection). One possible reason is that experts likely left out some related records that the agent needs in order to understand the evidence. This is evidence that even experts sometimes could not reflect the real work environment well and may unnecessarily make the tasks impossible or harder \citep{openai2024verified,zhu2025abc}.

We also observed that the differences between agents changed across environments. The three agents all scored higher in the task-specific environment compared to the full projection, but by different amounts. Grok 4.6 had an increase of 12.8, whereas Claude Opus 5 increased only 6.2. Initially, Opus surpassed Grok by 14.2 under the full projection, but in the task-specific environment this decreased to only 7.7. The highest-scoring agent also switched in three tasks. Under the two different environments, we had fixed the tasks and documents, and the only difference was that the agent needed to find the evidence itself. So we think under the full projection, the advantage of Opus compared to Grok largely comes from its ability in evidence finding, which is an important capability in real work. However, when we created the task-specific environments, this was stripped away.

Comparing agent performance on the two different environments two experts built separately for the same task, we found that the agent pass rate on average changed by 6.6 percentage points, which is almost 1/3 of the gap between the strongest and weakest agents (19.8 points). Other fields have also found similar things: estimates varied widely when 29 teams used the same data set to answer the same question, and this variation was not explained away by the analysts' expertise \citep{silberzahn2018}. In the field of machine learning, annotators often disagree on labels, and the disagreements are often real rather than noise \citep{plank2022}. This disagreement is amplified in agent environment building, where experts can make different but reasonable choices according to their personal experience and understanding of the work, and this difference can change the score substantially. This is also evidence for the call for more representative design of work, where we don't only need one or two experts but a board of experts to design construct-valid tasks (and environments).

Current benchmark checklists require builders to document how their data were made and check that tasks can be solved \citep{reuel2024betterbench,zhu2025abc}. To take this further and solve the problems we observed, we argue that benchmarks for agents should also report how the environment materials were selected, and if possible, test how much the score changes with a different builder. This is most important because it costs a lot of resources to build the environments and tasks and run experiments. In our study, one company with only eight tasks used most of the budget since experts and agents all had to read long documents. If there are no minimal reporting standards, large-scale evaluations will not be comparable or repeatable, so the cost would go to waste.

This study has a few limitations. First, our estimates came from only eight tasks from one pharmaceutical company, so the interval is wide for all pass differences. In pharmaceutical companies, there are abundant long documents, and experts and agents need to read a lot of documents and texts, which costs a lot of time and money. Experts spent roughly 110 hours writing tasks, mapping evidence, building task-specific environments, and checking grader scores and access labels (at an expert rate of \$120 per hour). The 192 runs were finished in five days, roughly 53 agent-hours. This study cost roughly \$18,000, excluding pilot tests and error adjusting (experts \textasciitilde\$13,000; agents and grader \textasciitilde\$5,000). Under our research budget, we were only allowed to build one company. We release the design of WorkWorlds and a reference implementation on a fictional company, so future studies could use this setting to test whether our findings hold in other companies and tasks.

Second, the environment was not interactive. Every run started from a fixed projection on a certain date. No one changed files in the environment during the run. This is intentional to keep the comparison clean and the results easy to interpret, similar to most static benchmarks \citep{xie2024osworld,xu2025theagentcompany}. Third, for the task-specific environments, experts only selected roughly 1\% of the files of the full projection. We did not intervene in this design, so we think this means task-specific environments constructed by experts are typically this small. We have no knowledge of the direction the effect would go when the environment builder selects more files. Finally, most criteria are graded by an LLM grader, GPT-5.6 Sol, which is the same as one of the tested models. Research has shown that graders tend to prefer outputs generated by the same models \citep{panickssery2024}. But since the main comparison in our study is the difference of the same agent on two environments, this bias should roughly cancel out. But it might affect the results we have on the differences between agents. Nonetheless, in our audit, the grader gave scores that agreed with the experts the most for GPT-5.6 Sol outputs (96.3\% vs. Grok 93.8\% and Opus 91.3\%), so we did not observe self-preference.

\section{Conclusion}

We built a company before the tasks and ran the same eight tasks under two different environments: actual environments that exist before the task vs. environments experts selected for the tasks. Expert-selected environments raised the pass rate from 68.0\% to 76.7\%, because agents found the evidence more often. When agents found the evidence, their pass rates were almost the same in the two environments. Switching between the two experts' environments for the same task changed the pass rate by 6.6 percentage points, and the three agents gained by different amounts, which changed the gaps between them. We argue that benchmarks should report how environment materials were selected and, if possible, they should test agent score changes after changing the environment builder.

\section*{AI Use Statement}

Generative AI tools (Claude Fable 5.1 High) were used for deep retrieval of relevant articles, assistance with writing and code, and generating abstracted code for release from the authors' existing code repository and infrastructure (proprietary). The paper was drafted, revised, and finalized by the authors.

\bibliography{refs}
\bibliographystyle{tmlr}

\appendix
\counterwithin*{table}{section}
\renewcommand{\thetable}{\thesection\arabic{table}}

\section{Construction Rules}\label{app:A}

Every evaluation follows three rules. First, tasks with the same revision, date, and employee seat start from the same workplace. Second, changing the seat only changes what the employee can see. Third, the rubric and scoring-only materials never enter the agent's environment. A run that breaks any rule is dropped. For each run, we keep the configuration, trajectory, workspace before and after the run, deliverables, and criterion-level results.

\section{World Instances}\label{app:B}

Besides the eight task-related areas, PharmaCo also contains products at other lifecycle stages and services unrelated to the tasks (e.g., calendars, internal messages, and other departments' records).

The three secondary worlds were built from public materials and frozen before a key event, so later materials stay on the verifier side. OncologyCo comes from an oncology drug company before its approval was made public (three regulatory writing tasks). DiagnosticsCo comes from the premarket approval (PMA) materials of a diagnostic device company before approval (four tasks, e.g., intended-use language and labeling). ClinicalSiteCo comes from public trial materials of a clinical site during site activation (four tasks, e.g., a site informed consent form and an institutional review board memo).

\section{Experts and Rubrics}\label{app:C}

All experts had at least ten years of relevant experience and were paid a fixed rate that did not depend on model performance or review results. Two domain experts wrote the tasks, checked the rubrics we wrote from the real events, and each built a task-specific environment for every task. Two other reviewers checked if each criterion had enough evidence in what the employee could see, and an adjudicator settled their disagreements. Separate domain experts did the grader audit, and nobody reviewed their own work.

We refined the criteria with several exploratory evaluations, in which the experts read outputs without knowing the model. These runs are not in the results. The rubrics, the reference notes used by the grader, and the grader settings were fixed before the measured runs.

\section{Construction Checks}\label{app:D}

Seat fingerprints cover what the agent can see (file contents, file dates, and tools), not container timestamps. Any exact file match between the verifier side and the agent's side was reviewed by hand. We found only one, a protocol the employee is meant to see that the grader also cites.

For evidence sufficiency, Reviewer 1 judged 67 criteria as having enough evidence and Reviewer 2 judged 69, and they agreed on 66 (91.7\%). Because nearly all criteria had enough evidence, Cohen's kappa was unstable (0.21, 95\% interval $-$0.05 to 0.42; \citealp{cohen1960}), so we also report Gwet's AC1 (0.91; \citealp{gwet2008}) and the overlap of the evidence locations they chose (mean Jaccard 0.77; \citealp{artstein2008}). Six criteria went to adjudication because the reviewers disagreed. The seventh went because both reviewers judged its evidence insufficient, and it was the one environment gap we repaired.

Table~\ref{tab:D0} summarizes the construction checks on PharmaCo reported in Section~\ref{sec:results}.

\begin{table}[H]
\centering
\caption{Construction checks on PharmaCo across eight tasks and 72 criteria.}
\label{tab:D0}
\small
\begin{tabular}{llr}
\toprule
\textbf{Property} & \textbf{Check} & \textbf{Result} \\
\midrule
Reproducibility & Clean rebuilds identical & 3/3 \\
 & Repeated seat instantiations identical & 18/18 \\
Seat isolation & Distinct seat fingerprints & 6/6 \\
 & Unauthorized exposure & 0 \\
 & Expected files present & 1,597/1,597 \\
Projection consistency & Tasks sharing a seat start from the same projection & 2/2 \\
 & Loaded workspace matches its seat & 8/8 \\
Verifier isolation & Planted errors caught before any run & 4/4 \\
 & Files on both candidate and verifier sides (reviewed) & 1 \\
Evidence sufficiency & Reviewer agreement & 66/72 \\
 & Criteria adjudicated & 7 \\
 & Environment gaps repaired & 1 \\
 & Mean evidence-location Jaccard & 0.77 \\
\bottomrule
\end{tabular}
\end{table}

Table~\ref{tab:D1} lists the 12 conformance checks and six planted violations on the public sample world, all of which passed or were caught.

\begin{table}[H]
\centering
\caption{Conformance and implant isolation checks for the sample world.}
\label{tab:D1}
\small
\begin{tabularx}{\textwidth}{L{0.33\textwidth}Xr}
\toprule
\textbf{Check} & \textbf{Exposed Failure} & \textbf{Result} \\
\midrule
\multicolumn{3}{l}{\textbf{Conformance Suite}} \\
Seat projection & Positions with different permissions receive identical evidence or incorrect projection hash & 4/4 \\
Date projection & Future records leaked, or date changed without projection change & 3/3 \\
Verifier sealing & Verifier-exclusive evidence enters candidate side, or required verifier materials missing & 2/2 \\
Explicit scoring path & Declared deterministic or semantic paths fail to produce expected judgment & 1/1 \\
Cross-surface consistency & Filesystem and record adapters expose differing paths or contents & 2/2 \\
\midrule
\multicolumn{3}{l}{\textbf{Implant Isolation Examples}} \\
Verifier-exclusive file copied to candidate package & Candidate/verifier isolation check & Detected \\
Approval records post-dating inspection copied to early projection & Date projection check & Detected \\
Quality-exclusive records copied to operations seat & Seat projection check & Detected \\
Candidate revision altered & Revision identity check & Detected \\
Projection hash replaced & Manifest integrity check & Detected \\
Path set of records surface altered & Cross-surface consistency check & Detected \\
\bottomrule
\end{tabularx}
\end{table}

\section{Environment Experiment Supplementary Results}\label{app:E}

\paragraph{Runs.} Table E5 lists the models, harnesses, timeouts, grader, and seeds. WorkWorlds set no limit on tool calls or context length. Thirteen attempts failed (five crashes, three out-of-memory errors, two provider errors, and three timeouts) and were run again. Each task and agent was run in two blocks of four (two full-projection runs and one run per expert environment).

\paragraph{Access labels.} A criterion counted as accessed only if the agent read every source in at least one sufficient evidence set. Seeing a path in a directory listing, search result, or filename did not count. When the trace could not decide this (e.g., a truncated trace), the label was unknown. A file counted as read only when the tool returned its content, and a record counted as read only when the tool returned the mapped record. A partial read counted only if the returned part contained the required content. Each unknown label recorded a reason: ambiguous source match (12), missing tool output (12), unsupported surface log (7), truncated trace (6), and other (1).

\paragraph{Audits.} Domain experts reviewed 240 criterion executions sampled by condition, task, and agent, without knowing the model, condition, expert, or grader judgment. They agreed with the grader on 93.8\% (Cohen's kappa = 0.85), 92.5\% under the full projection and 95.0\% under the task-specific environment. By agent, agreement was 91.3\% for Claude Opus 5, 93.8\% for Grok 4.6, and 96.3\% for GPT-5.6 Sol. Before labeling all data, two coders independently labeled a stratified sample of 346 criterion executions and agreed on 89.6\% (Cohen's kappa = 0.73, 95\% interval 0.64 to 0.80). Five criteria (120 executions) used deterministic scoring, and the other 67 (1,608 executions) used semantic scoring.

\begin{table}[H]
\centering
\caption{Raw results of the environment experiment for 72 criteria. Access rates use executions with observable access tags. Pass rates after finding evidence are descriptive only.}
\label{tab:E1}
\small
\begin{tabular}{lrr}
\toprule
\textbf{Result} & \textbf{Full projection} & \textbf{Task-specific environment} \\
\midrule
Criterion executions & 864 & 864 \\
Passed / Executed & 589/864 (68.2\%) & 664/864 (76.9\%) \\
Accessed / Observable & 614/842 (72.9\%) & 767/848 (90.4\%) \\
Unknown access tags & 22 (2.5\%) & 16 (1.9\%) \\
Pass rate after finding evidence & 84.2\% & 82.7\% \\
\midrule
\multicolumn{3}{l}{\textbf{Joint status of access and pass}} \\
Accessed, passed & 517 & 634 \\
Accessed, failed & 97 & 133 \\
Not accessed, passed & 53 & 18 \\
Not accessed, failed & 175 & 63 \\
\midrule
\multicolumn{3}{l}{\textbf{Search volume (mean per run)}} \\
Files/records viewed & 28.6 & 15.2 \\
Tool calls & 34.1 & 23.2 \\
Tool calls before first finding sufficient evidence & 12.1 & 7.8 \\
\bottomrule
\end{tabular}
\end{table}

Table~\ref{tab:E1} gives the raw counts behind the GEE estimates in the main text. Files and records viewed count every read in a run.

\begin{table}[H]
\centering
\caption{Results of the environment experiment by task. $N$ represents criterion executions per condition for each task. Access uses observable tags only; the final column reflects merged pass differences recalculated with each task omitted.}
\label{tab:E2}
\small
\begin{tabular}{lrrrrrr}
\toprule
\textbf{Task ($N$ / condition)} & \textbf{Full pass} & \textbf{Env pass} & \textbf{$\Delta$ pp} & \textbf{Full access} & \textbf{Env access} & \textbf{\shortstack{Merged $\Delta$ pp\\omitting task}} \\
\midrule
Stability update (132) & 64.4\% & 74.2\% & +9.8 & 76.2\% & 88.5\% & +8.5 \\
Overwrap position (132) & 68.9\% & 73.5\% & +4.5 & 76.0\% & 92.2\% & +9.4 \\
Validation defense (108) & 78.7\% & 65.7\% & $-$13.0 & 78.5\% & 78.5\% & +11.8 \\
Authority response (168) & 66.7\% & 78.0\% & +11.3 & 71.0\% & 95.8\% & +8.0 \\
Records integrity (108) & 66.7\% & 84.3\% & +17.6 & 73.8\% & 92.4\% & +7.4 \\
Supplier-change audit (96) & 59.4\% & 86.5\% & +27.1 & 57.4\% & 92.6\% & +6.4 \\
Shelf-life assessment (60) & 70.0\% & 73.3\% & +3.3 & 77.6\% & 88.1\% & +9.1 \\
Qualification sweep (60) & 75.0\% & 81.7\% & +6.7 & 72.9\% & 93.2\% & +8.8 \\
\bottomrule
\end{tabular}
\end{table}

Validation defense was the only task where the task-specific environment scored lower, and the only task where access did not change.

\begin{table}[H]
\centering
\caption{Results of the environment experiment by agent, with 288 criterion executions per agent per condition.}
\label{tab:E3}
\small
\begin{tabular}{lrrrr}
\toprule
\textbf{Agent} & \textbf{Full pass} & \textbf{Env pass} & \textbf{Pass $\Delta$ pp} & \textbf{Access $\Delta$ pp} \\
\midrule
Claude Opus 5 & 229/288 (79.5\%) & 247/288 (85.8\%) & +6.2 & +12.8 \\
Grok 4.6 & 188/288 (65.3\%) & 225/288 (78.1\%) & +12.8 & +17.5 \\
GPT-5.6 Sol & 172/288 (59.7\%) & 192/288 (66.7\%) & +6.9 & +22.3 \\
\bottomrule
\end{tabular}
\end{table}

GPT-5.6 Sol gained the most in access (+22.3 points) but only +6.9 in pass, whereas Grok 4.6 gained +17.5 in access and +12.8 in pass.

We also compared each expert's environments with the full projection separately (raw pass rates; 95\% intervals from 10,000 task bootstraps). Expert 1's environments were 9.1 percentage points above the full projection (2.3 to 14.7) and Expert 2's were 8.2 points above ($-$1.2 to 16.7). Within a task, the two experts' environments differed in pass rate by 6.6 points on average (4.2 to 9.2).

\begin{table}[H]
\centering
\caption{Task-specific environments built by the two experts. Counts are sources each expert selected (after review, where it changed). Jaccard compares the two experts' selections. The last column is the number of sources each expert added after the evidence check. Expert 2's first environment for authority response left out the Module 2.3 excerpt, which every one of its 14 criteria needs, so it covered none of the criteria until this one file was added.}
\label{tab:E4}
\small
\begin{tabular}{lrrrc}
\toprule
\textbf{Task} & \textbf{Expert 1} & \textbf{Expert 2} & \textbf{Jaccard} & \textbf{\shortstack{Added after check\\(Expert 1 / Expert 2)}} \\
\midrule
Stability update & 11 & 11 & 0.83 & 0 / 0 \\
Overwrap position & 10 & 10 & 0.82 & 0 / 0 \\
Validation defense & 9 & 8 (9) & 0.70 & 0 / 1 \\
Authority response & 9 & 9 (10) & 0.80 & 0 / 1 \\
Records integrity & 9 & 9 & 0.80 & 0 / 0 \\
Supplier-change audit & 9 & 9 & 1.00 & 0 / 0 \\
Shelf-life assessment & 7 & 7 & 1.00 & 0 / 0 \\
Qualification sweep & 8 & 8 & 1.00 & 0 / 0 \\
\bottomrule
\end{tabular}
\end{table}

In the three tasks with a Jaccard of 1.00, the two experts picked the same files.

\begin{table}[H]
\centering
\caption{Run settings and seeds for the environment experiment.}
\label{tab:E5}
\small
\begin{tabularx}{\textwidth}{L{0.27\textwidth}X}
\toprule
\textbf{Setting} & \textbf{Value} \\
\midrule
Claude Opus 5 & claude-opus-5-20260318; Claude Code; medium reasoning effort \\
Grok 4.6 & grok-4.6-20260312; Cursor CLI; high preset model configuration \\
GPT-5.6 Sol & gpt-5.6-sol-20260320; Codex; medium reasoning effort \\
Candidate timeout & 1,800 s for shelf-life assessment and qualification sweep; 3,600 s for the other six tasks \\
Tool call and context limits & None imposed by WorkWorlds \\
Semantic grader & gpt-5.6-sol; batch after all runs; blind to condition and expert \\
Run schedule & Two 4-run blocks per task and agent; each block has 2 full projection runs and 1 run per expert environment \\
Attempts & 205 attempts for 192 completed runs \\
Run order seed & 2026092193 \\
Bootstrap & 10,000 task-cluster replicates; seed 20270908 \\
Analysis software & numpy 2.3.5, pandas 2.2.3, scipy 1.17.0, statsmodels 0.14.6 \\
\bottomrule
\end{tabularx}
\end{table}

\section{Release}\label{app:F}

We release the de-identified criterion-level and run-level results, the analysis code, and a sample world with its conformance and isolation tests. We do not release PharmaCo, its source materials, scoring-only evidence, raw trajectories, or the production code.

\end{document}